\documentclass[letterpaper, 10 pt, conference, korean]{ieeeconf}  % Comment this line out if you need a4paper
\usepackage{kotex}
\usepackage{multirow}
\usepackage{subfigure}
\usepackage{bm}
\usepackage{euscript}
\usepackage{bbm}
\usepackage{booktabs} % for professional tables

\usepackage{hhline}
\usepackage{comment}
\usepackage{verbatim}
\usepackage{graphicx}
\usepackage{bbold}
\usepackage{amsmath}
\usepackage{xcolor}

\IEEEoverridecommandlockouts                              % This command is only needed if 
\title{\LARGE \bf
Self-supervised 3D Object Detection from\linebreak
Monocular Pseudo-LiDAR
}

\author{Curie Kim$^{1}$, Ue-Hwan Kim$^{2}$ and Jong-Hwan Kim$^{*}$% <-this % stops a space
\thanks{This work was supported in part by the Institute for Information \& communications Technology Promotion (IITP) grant funded by the Korea government (MSIT) (No.2020-0-00440, Development of Artificial Intelligence Technology that Continuously Improves Itself as the Situation Changes in the Real World, No.2022-0-00907, Development of AI Bots Collaboration Platform and Self-organizing AI) and the National Research Foundation of Korea (NRF) grant funded by the Korea government (MSIT) (No. NRF-2022R1C1C1009989).}
\thanks{$^{1}$Curie Kim is a research engineer in Samsung Electronics Co., Ltd., Suwon-si Gyeonggi-do, Korea
        {\tt\small curie3170@gmail.com}}%
\thanks{$^{2}$Ue-Hwan Kim is with Faculty of AI Graduate School, Gwang-ju Institute of Science and Technology (GIST), Gwang-ju, Korea 
        }%
\thanks{$^{*}$Jong-Hwan Kim is with Faculty of Electrical Engineering School, Korea Advanced Institute of Science and Technology (KAIST), Daejeon, Korea 
        }%        
}

\begin{document}

\maketitle
\thispagestyle{empty}
\pagestyle{empty}

%%%%%%%%%%%%%%%%%%%%%%%%%%%%%%%%%%%%%%%%%%%%%%%%%%%%%%%%%%%%%%%%%%%%%%%%%%%%%%%%
\begin{abstract}

There have been attempts to detect 3D objects by fusion of stereo camera images and LiDAR sensor data or using LiDAR for pre-training and only monocular images for testing, but there have been less attempts to use only monocular image sequences due to low accuracy. In addition, when depth prediction using only monocular images, only scale-inconsistent depth can be predicted, which is the reason why researchers are reluctant to use monocular images alone. Therefore, we propose a method for predicting absolute depth and detecting 3D objects using only monocular image sequences by enabling end-to-end learning of detection networks and depth prediction networks. As a result, the proposed method surpasses other existing methods in performance on the KITTI 3D dataset. Even when monocular image and 3D LiDAR are used together during training in an attempt to improve performance, ours exhibit is the best performance compared to other methods using the same input. In addition, end-to-end learning not only improves depth prediction performance, but also enables absolute depth prediction, because our network utilizes the fact that the size of a 3D object such as a car is determined by the approximate size.
\end{abstract}

%%%%%%%%%%%%%%%%%%%%%%%%%%%%%%%%%%%%%%%%%%%%%%%%%%%%%%%%%%%%%%%%%%%%%%%%%%%%%%%%

\section{INTRODUCTION}
\iffalse The detailed technologies constituting autonomous driving include control technology, motion planning, path discovery, and obstacle avoidance autonomy, and all detailed technologies operate based on the environment recognized by the sensor. \fi Recognizing geometric features as well as semantic entities of the surrounding environment is essential for intelligent agents to output meaningful information  \cite{9712359}. Since the most important thing for detailed technology operation is to detect the position and state of an object in 3D, sensors that measure depth information, such as a stereo camera, RGBD camera, and 3D LiDAR sensor, are essential \cite{kim2019simvodis}.

\begin{figure}[t]
    \centerline{\includegraphics[width=8.5cm]{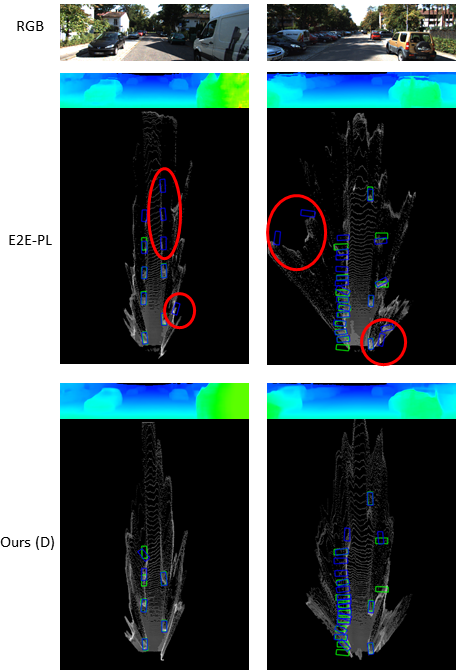}}
    \caption[overview]{Depth estimation results and 3D object detection results of E2E-PL \cite{Qian_2020_CVPR} and Ours (D) to the corresponding RGB images. False positives incorrectly detected by E2E-PL of cars on the road are circled in \textcolor[rgb]{1,0,0}{red}. \textcolor[rgb]{0,1,0}{Green} - correct answer label, \textcolor[rgb]{0,0,1}{blue} - predicted result.
    } \label{overview-fig}
\end{figure}

However, these sensors are expensive, and the amount of data sets on the market is absolutely insufficient. Moreover, contrary to the expectation that the 3D object recognition performance would be improved by using the image and LiDAR data together, the performance when using the LiDAR alone is much better. Therefore, the barriers to overcome for 3D object recognition for autonomous driving are: 1) the high price of high-performance 3D sensors, 2) the absence of prior research on 3D object detection, and 3) fusion is impossible due to the gap between the 2D sensor and the 3D sensor.

The goals of this study are as follows. First, we propose a monocular image-based algorithm to reduce the cost of the sensor and provide extensibility to experiment with larger datasets. In PL \cite{Wang_2019_CVPR}, also a method of using a monocular image as an input was suggested, but it is not possible to perform end-to-end learning. Mono PL  \cite{weng2019monocular} capable of end-to-end learning was developed, but in either case, it cannot be said that only monocular images are used perfectly. PL and Mono PL use DORN \cite{fu2018deep} as a depth estimation method because DORN networks have already used LiDAR information for pre-training. However, since our proposed depth estimation network uses self-supervised loss, it is possible to learn from only monocular images without LiDAR data even during pre-training and end-to-end learning. Second, we propose a method to improve the mono-based 3D object detection performance. Only monocular images are used as input, but additional experiments to obtain supervised loss from LiDAR are also conducted to achieve state-of-the-art performance.

Moreover, we propose three loss calculation methods. The first one is a method of calculating unsupervised loss from monocular images of three sequential frames (M). In this case, only moncular image input is used, and learning is possible without using LiDAR data in any case of pre-training. The second one is a method that uses both images and LiDAR (D). The supervised loss can be calculated from LiDAR. The last one is a method that uses all of the aforementioned losses (MD).

\iffalse The pipeline we designed is shown in Fig. \ref{overview-fig}. Our method is based on the previously introduced PL concept, which contributes to the improvement of image-based algorithm performance by first changing the expression method of depth data by filling the sparse LiDAR with a dense 3D point cloud. We perform mono depth estimation using Monodepth2 \cite{godard2019digging}, project the estimated depth into 3D, convert it to Pseudo-LiDAR, and then convert the generated pseudo-LiDAR to PIXOR  \cite{yang2019pixor}, a 3D object detection algorithm. This entire training process is an end-to-end learning. In the process of projecting the estimated depth in 3D, there was a problem that back propagation did not occur. This was solved by using the Change of Representation (CoR) method proposed in E2E-PL \cite{Qian_2020_CVPR}. \fi

As a result, we significantly reduce the performance difference between mono-based and stereo-based algorithms. Among the models proposed by us, the self-supervised model (M) exhibits SOTA performance exceeding the existing mono-based algorithms \cite{MonoGRNet, MonoPair, Monodle, GUPNet, Homo} in the car category of the KITTI 3D validation dataset, assuring a large performance gap especially when IoU = 0.5. The supervised model (D) records SOTA performance in the moderate and hard difficulty levels in the car category of the KITTI 3D validation dataset. \iffalse When evaluating moderate and hard difficulty in the car category at IoU = 0.5, 7.13\%, 9.94\%, and IoU = 2.75\%, 4.63\%  $AP_{BEV}$ were improved compared to Mono PL \cite{weng2019monocular}, respectively. \fi

In addition, the proposed method enables absolute depth estimation from monocular images. In the case of the conventional mono depth estimation, only the relative disparity could be estimated. \iffalse The depth map estimated by the monocular image seemed to be well estimated in 2D, but when projected in 3D, it was not accurate. Attempts to detect 3D objects with this inaccurate 3D point cloud resulted in low accuracy. \fi However, since our method learns the object detection loss in 3D along with the depth estimation network, we are able to estimate the absolute depth on a consistent scale by learning the geometrical assumption. Mono PL\cite{weng2019monocular} places additional limitations to solve the problem of local misalignment of PL and long-tail of the point cloud of object boundaries. But, our algorithm, through end-to-end learning, allows the network to estimate the depth at which it naturally leads to a 3D object detection. \iffalse The RMSE of the existing Monodepth \cite{godard2019digging} was 4.863m, and it can be seen from the decrease in our depth estimation result to 4.415m. \fi

In summary, the contribution of this paper is as follows.
\begin{itemize}
    \item We propose a method to train with self-supervised loss using only monocular image sequences, which  outperforms previous algorithms by more than 3\% $AP_{BEV}$ at IoU = 0.5.
    \item In the case of fusing monocular image sequences and LiDAR together during training, we record SOTA performance in moderate and hard difficulty levels in the car category of the KITTI 3D validation dataset.
    \item Through end-to-end learning, when estimating depth from monocular image, it is possible to predict absolute depth, which was previously impossible.
    \item We narrow the performance gap between mono-based and stereo-based methods in detecting 3D objects.
\end{itemize}

\iffalse
\cite{Wang_2019_CVPR} proposed a new concept of Pseudo-LiDAR (PL), arguing that the reason for the difficulty of fusion between 2D images and LiDAR is the expression method of LiDAR. PL fuse the sensors by filling the existing sparse LiDAR with a high density using a depth estimation technique and projecting it in 2D. Later, E2E-PL\cite{Qian_2020_CVPR}, a technique for end-to-end learning of PL, was developed to reduce the performance gap between image-based and LiDAR-based 3D object detection. However, it still costs the 3D sensor as it uses stereo images as input and requires LiDAR for supervised learning. However, 3D object detection in mono images is an integral part of future autonomous driving \cite{kiran2020deep} and robotic vision \cite{manglik2020forecasting}, as inexpensive onboard cameras are readily available in most modern cars. Unfortunately, the performance of the existing mono image-based methods is very low compared to the performance of the LiDAR-based and stereo-based methods. The average performance of the mono image-based method \cite{8578347} is 13.6\% AP (average precision) when detecting the car category of medium difficulty in the KITTI 3D dataset  \cite{geiger2013vision}, whereas the average performance of the LiDAR-based method  \cite{ku2018joint} is 86.5\% AP, which is significantly different \cite{weng2019monocular}.
\fi
\label{Related Works}
\section{Related Works}
\subsection{Depth Estimation}

\textbf{Supervised Depth Estimation.}
Eigen \cite{eigen2014depth} first developed a network that can estimate depth from monocular images without special preprocessing. By proposing a coarse-to-fine network, it first extracts global features to predict the depth, and then improves the depth more precisely. \iffalse GC-Net \cite{kendall2017endtoend} is the first to propose a 3D cost volume to predict disparity from stereo images. End-to-end learning is possible by predicting the disparity by performing soft argmin operation on the 3D cost volume, and learning with supervised regression loss using LiDAR disparity information. PSMNet  \cite{chang2018pyramid} improved performance by using Spatial Pyramid Pooling (SPP), which was effective in segmentation, for depth prediction as well. PSMNet covers various receptive fields and use the 3D cost volume loss proposed by GC-Net. \fi DORN  \cite{fu2018deep} learns by transforming depth information into ordinal regression through spacing-increasing discretization instead of learning with linear regression when estimating monocular depth.

\textbf{Monocular Unsupervised Depth Estimation.}
Monodepth \cite{godard2017unsupervised} estimates the depth by unsupervised learning by defining the image reconstruction loss to be the same as the original opposite image when projecting one stereo image to the opposite image with the predicted disparity. \iffalse Monodepth calculates the image reprojection loss by knowing the camera pose, but SfmLearner \cite{8100183} reprojects the image to the camera pose obtained by training a network to predict the camera pose between two sequential images. A camera pose prediction network and a depth estimation network have been designed, which require the assumption that objects in the image are not moving. Deep-VO-Feat \cite{zhan2018unsupervised} uses the difference between features as a loss from the idea that not only the image and the reprojected image should be the same, but the convolutional features of those images should be the same. Struct2Depth\cite{casser2018depth} uses a mask using instance segmentation information to predict the direction and speed of a moving object in the image to improve performance.\fi Monodepth2 \cite{godard2019digging} learns the camera pose through a separate network and improves the loss used in the existing Monodepth. \iffalse When calculating the reprojection loss, the average was previously used, but the minimum value is used to make it more robust to occluded pixels. An auto-masking technique is also used to remove the hole that occurred when the moving object is reprojected. Manydepth \cite{manydepth} challenges to the problem that monocular unsupervised depth estimation methods cannot predict the exact scale of depth.\fi

\iffalse
\textbf{Stereo Unsupervised Depth Estimation.}
UnDeepVO \cite{li2018undeepvo} also predicts the camera pose similar to SfmLearner's unsupervised loss described above. Unlike SfmLearner, it uses stereo images to know the absolute scale of depth. GeoNet  \cite{yin2018geonet} predicts the optical flow as well as the depth and camera pose. It learns by calculating the left-eye coherence loss from the predicted optical flow.
\fi

\subsection{3D Object Detection}
\textbf{LiDAR-based Methods.}
Convolutional Neural Networks (CNNs) have been developed to process 2D images, and are not suitable for processing sparse 3D LiDAR data. \iffalse Therefore, VoxelNet \cite{zhou2017voxelnet}, SECOND \cite{yan2018second} designs and applies a 3D CNN to process data in sparse 3D space.\fi Alternatively, to utilize the existing 2D CNN, PIXOR \cite{yang2019pixor} projects the point cloud in two dimensions and processes it. \iffalse PointRCNN \cite{shi2019pointrcnn} detects 3D objects from raw point clouds.  \fi

\textbf{Image-based Methods.}
\iffalse  Deep3Dbox \cite{mousavian20173d} solves optimization problems and develops 2D bounding boxes into 3D by exploiting the geometric constraints that 3D boxes have inevitably due to 2D-3D consistency. However, this method has the disadvantage of being overly dependent on the 2D bounding box. \fi Multi-Level Fusion (MLF) \cite{xu2018multi} creates a point cloud by projecting RGB image pixels in 3D with pre-trained weights of Monodepth. The point cloud is then fused with RGB features to detect the 3D objects. Pseudo-LiDAR \cite{Wang_2019_CVPR} developed MLF, which generates a point cloud with DORN and then passes it through the highest performing LiDAR-based 3D object detection algorithm. Since then, many pseudo-LiDAR-based studies have been derived.

\textbf{Sensor Fusion Methods.}
\iffalse Frustum-PointNet \cite{qi2018frustum} and Frustum-ConvNet \cite{wang2019frustum} detect an object in a 2D image and limit the search area using only the 3D point cloud of the corresponding frustum. AVOD \cite{ku2018joint}, MV3D \cite{chen2017multiview} processes RGB images and LiDAR with different backbone networks, and then fusions the features of the two sensors in a Region Proposal Network (RPN), using Faster-RCNN \cite{ren2015faster} as a detector. ContFuse \cite{liang2018deep} is designed to fusion for each layer when image and LiDAR features are extracted, but the problem of feature blurring occurs because the resolution between the two data is different.\fi SemanticVoxels \cite{9235240} augment the point cloud by semantic features extracted from images. MonoDTR \cite{MonoDTR} adapt depth-aware transformer network \cite{Attention} while fusing LiDAR data and image features.

\section{Methodology}
\begin{figure*}[t]
  \centering{
  \includegraphics[width=16cm]{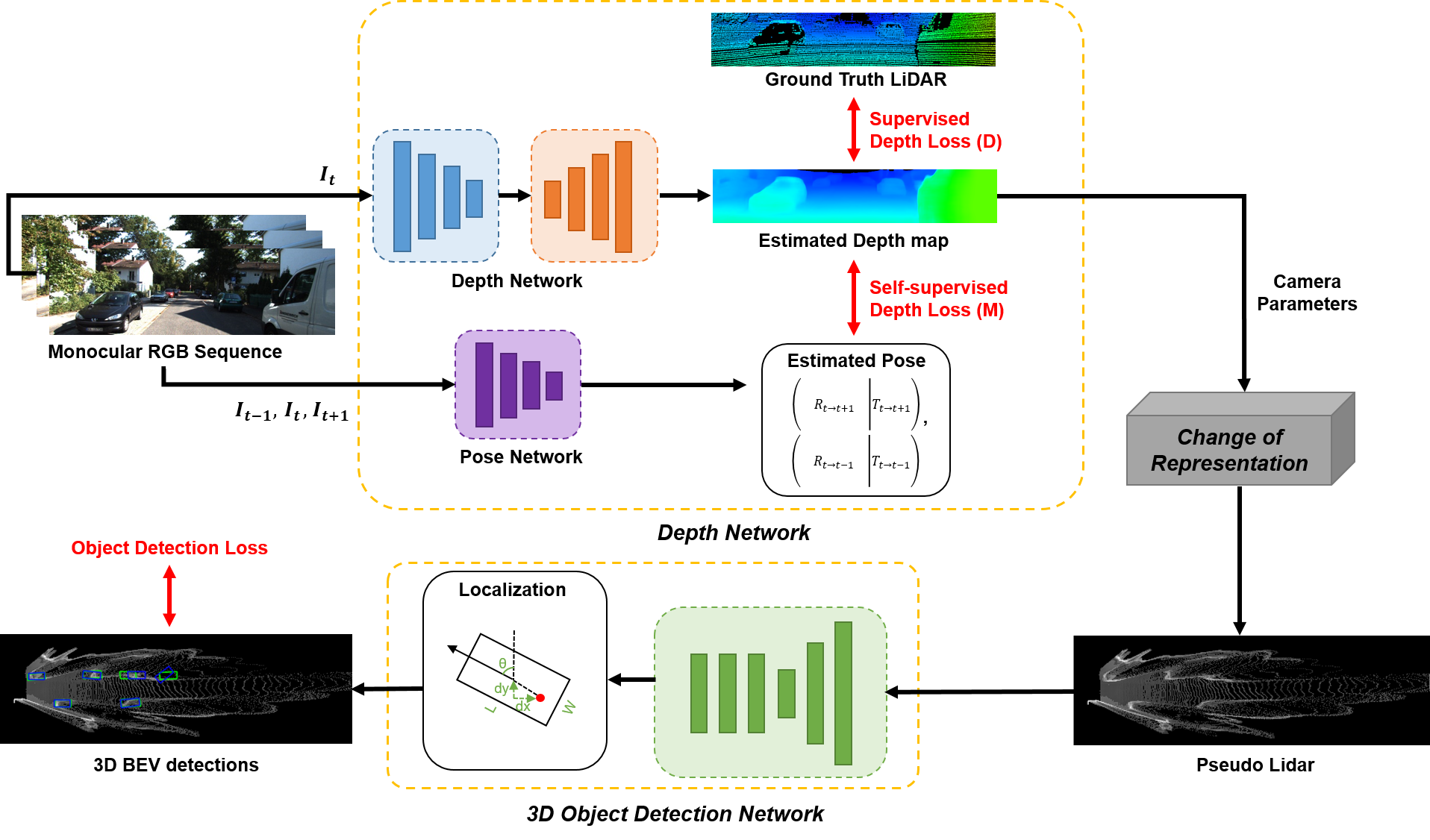}}
  \caption{\textbf{Our 3D object detection network.} Three sequential images, $\bm{I}_{t-1}$, $\bm{I}_{t}$, $\bm{I}_{t+1}$ are used as inputs to estimate the camera pose, while the depth network feeds only $\bm{I}_{t}$. Learning with supervised loss (D) or self-supervised loss (M) or both (MD) are available, and the predicted depth is converted into a pseudo-LiDAR from through a change of representation scheme proposed by \cite{Qian_2020_CVPR}. Then the 3D object network detects 3D objects by considering it as a LiDAR sensor measurement result.}
  \label{network-fig}
\end{figure*}
\subsection{Overview}

Fig. \ref{network-fig} displays the overall architecture of the proposed monocular 3D object detection framework. The framework consists of three main components: 1) monocular depth estimation, 2) PL point cloud generation, and 3) 3D object detection. First, the input monocular images pass through the monocular depth estimation module. For both supervised and self-supervised learning settings, only the center image ($\bm{I}_{t}$) gets fed into the depth network and becomes an estimated depth map. For self-supervised learning, the pose network receives two other temporal frames ($\bm{I}_{t-1}$ and $\bm{I}_{t+1}$) and estimates the relative camera motion. The depth estimation module can learn the estimation with supervision signals from geometric constraints (M) or ground-truth depths (D) or both (MD). Next, the PL point cloud generation module converts the estimated depth map into a PL point cloud in a way that gradients can propagate through the PL cloud. According to the 3D object detector that follows, the PL point cloud further gets either quantized (voxelized) or subsampled. Finally, the 3D object detection module performs 3D object detection on the PL point cloud.

\subsection{Scale-aware Depth Estimation}

Due to the inherent scale ambiguity of monocular depth estimation, the process of monocular 3D object detection could become unstable. To deal with this, we propose a scale-aware depth estimation method. The key to overcoming the scale ambiguity is to represent depths as follows:
\begin{equation}
    \hat{d} = \frac{\bar{D}_{\text{prior}}}{\sigma_\text{min} + (\sigma_\text{max} - \sigma_\text{min}) \cdot x},
\end{equation}
where $x \in [0, 1)$ denotes the output of the depth network (disparity), $\sigma_\text{max}$ and $\sigma_\text{min}$ are the maximum and minimum values disparity can take, respectively and $\bar{D}_{\text{prior}}$ represents a scaling factor. We scale the output of the depth network twice. The first scaling places the estimated depths inside the range of $(1/\sigma_{\text{max}}, 1/\sigma_{\text{min}}]$. This scaling effectively makes the estimated depth scale-consistent over multiple scenes \cite{godard2019digging}. Next, we incorporate prior knowledge into the depth estimation process in the form of the scaling factor ($\bar{D}_{\text{prior}}$). An appropriate choice of the scaling factor lets the estimated depths lie in a reasonable range when combined with the scale-consistent depth estimation. We can set the scaling factor either by an empirical study or by an inductive process.

\subsection{Change of Representation}
\textbf{PL Point Cloud. } After the depth estimation module, each pixel coordinate $(u, v)$ with depth $d$ in the 2D image space becomes a point $\bm{p}=(x, y, z)$ in the 3D space as follows:
\begin{equation}
\begin{cases}
z &= d,\\
x &= (u - C_x) \cdot z/f,\\
y &= (v - C_y) \cdot z/f,
\end{cases}
\end{equation}
where $(C_x, C_y)$ is the camera center and $f$ is the focal length. The resulting point cloud ($\bm{P}$) contains all the generated points: $P = \{p_1, ..., p_N\}$, where $N$ is the total number of generated points.

\textbf{Soft Quantization. } Quantization aims to voxelize the region of interest and represent it as a 3D or 3D tensors. Since hard quantization hardly propagates gradients, we adopt the soft quantization technique \cite{Qian_2020_CVPR}. Soft quantization defines a 3D occupation tensor $\bm{T}$ of $M$ bins as follows:
\begin{equation}
    \bm{T}(m) = T(m, m) + \frac{1}{|\EuScript{N}_{m}|}\sum_{m' \in \EuScript{N}_{m}} \bm{T}(m, m'),
\end{equation}
where $m \in \{1, ..., M\}$ is a bin in $\bm{T}$, $\EuScript{N}_{m}$ represents the set of neighbor bins of $m$ and
\begin{equation}
    \bm{T}(m, m') = \begin{cases}
    0 & \text{if} \,\, |\bm{P}_{m'}| = 0,\\
    \frac{1}{|\bm{P}_{m'}|} \sum_{\bm{p} \in \bm{P}_{m'}} e^{-\frac{\Vert\bm{p} - \bm{\hat{p}}_m\Vert^2}{\sigma^2}} & \text{if} \,\, |\bm{P}_{m'}| > 0,
    \end{cases}
\end{equation}
where $\bm{\hat{p}}_m$ is the center of a bin $m$, $\bm{P}_{m} = \{\bm{p} \in \bm{P}, \text{s.t.}\,\, m=\underset{m}{\mathrm{argmin}}|| \bm{p} - \hat{\bm{p}}_m' ||^2\}$ is the set of points inside the bin $m$.

\subsection{Loss Functions}
\textbf{Depth Smoothness Loss.} We impose the depth smoothness loss on the estimated depth map to restrain random sharp lines \cite{mahjourian2018unsupervised}. The smoothness loss regularizes the gradients of the estimated depth map ($\bm{\hat{D}}_t$). Moreover, we penalize the smoothness loss by inversely weighting it with the gradients of the input monocular image so that essential gradients would remain. Finally, the smoothness loss is
\begin{equation}
    L_{smooth} = \sum|\partial_x \hat{\bm{D}}_t| \cdot e^{-|\partial_x \bm{I}_x|} + |\partial_y \hat{\bm{D}}_t| \cdot e^{-|\partial_y \bm{I}_y|}.
\end{equation}

\textbf{Supervised Depth Loss (D).} LiDAR signals provide direct supervision signals for depth learning. We formulate a $L_1$ loss for depth supervision by projecting 3D LiDAR signals to the 2D image space and comparing the projected depth map ($\bm{D}_t$) and the estimated depth map ($\bm{\hat{D}}_t$). In addition, we filter out the loss with a mask representing the pixels where a LiDAR signal exists since LiDAR signals are sparse. Finally, we define the loss as follows:
\begin{equation}
    L_D = \Vert \mathbbm{1}_{\{\bm{D}_t > 0\}} \cdot (\bm{D}_t - \hat{\bm{D}}_t) \Vert_1,
\end{equation}
where $\mathbbm{1}_S$ is an indicator function for a set $S$.

\textbf{Self-Supervised Depth Loss (M).} The photometric consistency loss \cite{zhou2017unsupervised} takes a central role in training the depth network with unlabeled temporal images. The loss restores the center image ($\bm{I}_{t}$) from two nearby views ($\bm{I}_{t-1}$ and $\bm{I}_{t+1}$) using the estimated motion vectors ($\hat{\bm{T}}$) and the predicted depth map ($\hat{\bm{D}}_{t}$); enforces photometric consistency by minimizing the photometric error. For the reconstruction, we project each pixel in $\bm{I}_{t}$ onto the nearby views as follows:
\begin{equation}
    p_{n \rightarrow t} = \hat{\bm{K}} \cdot \hat{\bm{T}}_{t \rightarrow n} \cdot \hat{\bm{D}_t}(p_t) \cdot \hat{\bm{K}}^{-1} \cdot p_t,
\end{equation}
where $n \in \{t-1, t+1\}$, $\hat{\bm{K}}$ is the estimated camera intrinsic matrix, $\hat{\bm{T}}$ is an element in the special Euclidean group SE(3) consisting of a rotation matrix $\bm{R} \in \mathbb{R}^{3\times 3}$ and a translation vector $\bm{t} \in \mathbb{R}^{3}$) and $p_t$ represents the homogeneous coordinate of each pixel in $\bm{I}_t$. Finally, we evaluate the photometric consistency loss as $pe (\bm{I}_t, \bm{\hat{I}}_{n \rightarrow t})$, where
\begin{equation}
    pe(\bm{I}_a, \bm{I}_b) = \frac{\alpha}{2}(1 - SSIM(\bm{I}_a, \bm{I}_b)) + (1 - \alpha) \Vert \bm{I}_a - \bm{I}_b \Vert_1,
\end{equation}
where $\alpha=0.85$ and SSIM is the structural similarity measure \cite{wang2004image}.

\iffalse
Finally, the photometric consistency loss is
\begin{equation}
    L_{\text{photo}} = \sum_{p} |\hat{\bm{I}}_{recon, n}(p) - \bm{I}_t(p)|,
\end{equation}
where $p$ is the set of pixel coordinates.

\textbf{Structural Similarity (SSIM) Loss. } The SSIM loss overcomes the Lambertian assumption that the phonometric consistency loss implicitly makes and better handles complex illumination changes \cite{godard2017unsupervised}. SSIM calculates the similarity between two image patches ($a$ and $b$) as follows:
\begin{equation}
    SSIM(a,b) = \frac{(2\mu_a \mu_b+c_1)(2\mu_{ab}+c_2)}{(\mu_a^2+\mu_b^2+c_1)(\sigma_a+\sigma_b+c_2)},
\end{equation}
where $\mu$ and $\sigma$ are patch mean and variance, respectively, $c_1 = 0.01^2$ and $c_2 = 0.03^2$. For the SSIM loss, we extract $3 \times 3$ image patches from $\bm{I}_t$ and $\hat{\bm{I}}_{recon, n}$ and minimize
\begin{equation}
    L_{SSIM} = \sum_p 1-SSIM(s(\hat{\bm{I}}_{recon, n},p),s(\bm{I}_t,p)),
\end{equation}
where $s(\bm{I},\, p)=\{\bm{I}(p + i, p + j)|i,j\in\{-1, 0, 1\} \}$ samples a $3 \times 3$ patch centered at $p$ from an image $\bm{I}$.
\fi

\textbf{Combined Depth Loss (MD).} When using both the supervised and self-supervised depth losses, we make the two losses complementary to each other using two masks as follows:
\begin{equation}
    L_{MD} = \lambda_M (\mathbbm{1}_{\{L_D = 0\}} \cdot L_M)+ \lambda_D (\mathbbm{1}_{\{L_D > 0\}} \cdot L_D),
\end{equation}
where $\lambda_M$ and $\lambda_D$ are weighting factors.

\textbf{3D Detection Loss.} We utilize two loss functions for learning 3D object detection: the focal loss for classification and the smooth L1 loss for 3D bounding box regression. The focal loss improves the learning efficiency when there exists class imbalance in the training data; the point clouds for 3D object detection in general contain more points belonging to background than objects. The smooth L1 loss prevents outliers from deteriorating gradients and stabilizes the regression learning.

\section{Experiments}

\subsection{Settings}
\textbf{Dataset.}  We evaluate the performance of 3D object detection on the challenging KITTI 3D object detection benchmark \cite{geiger2013vision}. The KITTI 3D benchmark consists of 7,481 images with ground-truth 3D labels. The ground-truth labels comprise of three categories: easy, moderate and hard. We split the data into 3,712 training and 3,769 validation images. \iffalse In addition, KITTI 3D provides 64-beam Velodyne LiDAR point clouds, stereo images and camera calibration matrices corresponding to the 7,481 images. \fi
In our algorithm, an image sequence is required to obtain unsupervised loss, but the KITTI 3D dataset does not provide continuous images. The previous and next frames of the KITTI 3D image are obtained from the KITTI raw dataset. In this process, if the KITTI 3D image is the first frame or the last frame of the video sequence, it is not included in the training and testing because there are no previous and subsequent frames. Excluded images are 7 in the training set and 8 in the validation set, which are not expected to significantly affect performance.

\begin{table*}[t]
\renewcommand{\arraystretch}{1.2}
\caption[Comparison of the car category on the KITTI validation set]{Comparison of the car category on the KITTI validation set with existing mono-based studies.}
\label{tb:result_comparative}
\begin{center}
%\begin{sc}
\begin{tabular}{|l|c|c|ccc|ccc|}
\hline 
\multicolumn{1}{|c}{\multirow{2}[2]{*}{\textbf{Method}}} & \multicolumn{1}{|c}{\multirow{2}[2]{*}{\textbf{Reference}}} & \multicolumn{1}{|c}{{\textbf{Input }}} & \multicolumn{3}{|c}{$\textbf{AP}_\textbf{BEV}$ , \textbf{IoU=0.5}} & \multicolumn{3}{|c|}{$\textbf{AP}_\textbf{BEV}$ , \textbf{IoU=0.7}}\\
\cline{4-6}
\cline{7-9}
&&(+ pretrained input) & Easy & Moderate & Hard & Easy & Moderate & Hard\\
\hline\hhline{|=|=|=|===|===|}
\iffalse
Mono3D\cite{7780605} & & Mono & 30.5 & 22.4  &19.2   &5.2   & 5.2  &4.1  \\
Deep3DBox\cite{mousavian20173d}&& Mono & 30.0  & 23.8  & 18.8  &10.0   &7.7   & 5.3\\
MLF-MONO\cite{8578347}&& Mono &55.0 &  36.7 & 31.3  &22.0 &  13.6  &  11.6 \\
ROI-10D\cite{manhardt2019roi10d}&& Mono & 46.9  & 34.1 &30.5 & 14.5 & 9.9  & 8.7  \\
MonoFlex & Mono&   &   &  & &  & \\
AutoShape\cite{Autoshape}&& Mono, CAD & 67.66 & 49.74 & 42.71 & 28.89 & 21.11 & 17.72\\
MonoGround\cite{Monoground}&& Mono, Plane& 67.36 & 51.83 & 46.65 & 32.68 & 24.79 & 20.56\\
\fi
\hline
MonoGRNet\cite{MonoGRNet}&AAAI 2019& Mono & 54.21 & 39.69 & 33.06 & 24.97 & 19.44 & 16.30\\
MonoPair\cite{MonoPair}&CVPR 2020& Mono & 61.06 & 47.63 & 41.92 & 24.12 & 18.17 & 15.76\\
Monodle\cite{Monodle}&CVPR 2021& Mono & 59.78 & 45.84 & 41.60 & 23.97 & 19.23 & 16.70\\
GUPNet\cite{GUPNet}&ICCV 2021& Mono & 61.78 & 47.06 & 40.88 & \textbf{31.07} & 22.94& 19.75\\
Homo\cite{Homo}& CVPR 2022& Mono &-&-&-&31.04&22.99&19.84\\
\textcolor[rgb]{0,0,1}{Ours(M)} &-& \textcolor[rgb]{0,0,1}{Mono} & \textcolor[rgb]{0,0,1}{\textbf{64.35}}& \textcolor[rgb]{0,0,1}{\textbf{50.39}}&\textcolor[rgb]{0,0,1}{\textbf{44.03}}& \textcolor[rgb]{0,0,1}{28.25} &\textcolor[rgb]{0,0,1}{\textbf{23.19}}&\textcolor[rgb]{0,0,1}{\textbf{21.01}}\\
\hline
PL-MONO\cite{Wang_2019_CVPR}&CVPR 2019 &Mono(+LiDAR) &70.8   & 49.4  &42.7  & 40.6  & 26.3  & 22.9 \\ 
Mono PL\cite{weng2019monocular} &ICCVW 2019& Mono(+LiDAR)&  \textbf{72.1}  &53.1   & 44.6 & \textbf{41.9}&28.3  &24.5  \\
DD3D\cite{DD3D}& ICCV 2021&Mono, Depth& - & -  &-  &37.0 &29.4  &25.4 \\
MonoDTR\cite{MonoDTR} &CVPR 2022& Mono, LiDAR&  69.04  &52.47   & 45.90 & 33.33&25.35  &21.68  \\
\textcolor[rgb]{0,0,1}{Ours(D)} &-& \textcolor[rgb]{0,0,1}{Mono, LiDAR} &\textcolor[rgb]{0,0,1}{70.86}& \textcolor[rgb]{0,0,1}{\textbf{60.23}}&\textcolor[rgb]{0,0,1}{\textbf{54.54}}& \textcolor[rgb]{0,0,1}{37.42}&\textcolor[rgb]{0,0,1}{\textbf{31.05}} &\textcolor[rgb]{0,0,1}{\textbf{29.13}}\\
\hline
\end{tabular}
\end{center}
\end{table*}  
\subsection{Implementation Details}

The pre-trained parameters provided by Monodepth2 are used for the depth estimation network, and the resolution of the input and output is set to 640 $\times$ 192. In the case of end-to-end learning, unlike the existing Monodepth2, multi-scale learning by adjusting the input image to various sizes is not performed, and a resolution of 308 $\times$ 1248 is used.

To generate the input of the 3D object detection network, the scale-aware depth map, which is the result of the depth estimation network, is converted into a pseudo-LiDAR using soft quantization. The existing PIXOR network is borrowed for 3D object detection. Since PIXOR converts the 3D LiDAR point cloud into a bird's eye view (BEV) expression viewed from above, the generated Pseudo-LiDAR is converted to the BEV format and then passed through the PIXOR network. 

End-to-end learning is implemented with PyTorch, and most hyperparameters followed E2E-PL. Adam Optimizer is used, and the learning rate is multiplied by 0.1 every 10 steps from 0.001. For a total of 15 epoch training, the supervised learning model (D), the self-supervised learning model (M), and the model using both (MD) using 4 Titan Xp with a batch size of 6 took 5.5, 6, and 7 hours, respectively.

\subsection{Results and Analysis}
The main results are summarized in Table \ref{tb:result_comparative}. Our model (M), trained self-supervised using only monocular images, performed the best in the car category of all difficulty levels, and our model (D), using monocular images and 3D LiDAR together, performed in the car category of medium and hard difficulty levels. The highest $AP_{BEV}$ performance is achieved in both IoU = 0.5 and IoU = 0.7. This result confirms the possibility to predict the depth consistent with the scale.
 
\textbf{Effect of Depth loss.}
The results of the depth prediction experiments for the three loss types are shown in Table \ref{tb:ablation depth}. The depth estimation evaluation method followed \cite{eigen2014depth}, and the validation set of KITTI 3D \cite{geiger2013vision} data is evaluated. MD2 stands for Monodepth2 \cite{godard2019digging}, PIXOR$^*$ denotes PIXOR \cite{yang2019pixor} with soft quantization applied, and E2E stands for end-to-end learning. The performance of the end-to-end trained model using MD and D losses is better than the result of learning only with Monodepth2, and again, the model (D) using 3D LiDAR exhibited the best results.

\textbf{Scale-aware Depth Estimation.}
Table \ref{tb:ablation AP} shows 3D object detection performance according to end-to-end learning and three loss types. The performance of the existing PIXOR \cite{yang2019pixor} using 3D LiDAR is the highest. After learning Monodepth2 and PIXOR individually, the result of simply connecting them is not so good. In Table \ref{tb:ablation depth},  we compare the performance with and without adopting scale-aware depth estimation. This shows that inconsistency exists in the depth predicted by the prediction model, and it is very far from the actual depth. However, when end-to-end learning is performed only with unsupervised loss (M), the performance increased nearly 7 times, showing the effect of end-to-end learning. The 3D object detection result also showed the highest performance in the model (D) using 3D LiDAR.

\begin{figure}[t]
    \centerline{\includegraphics[width=8.5cm]{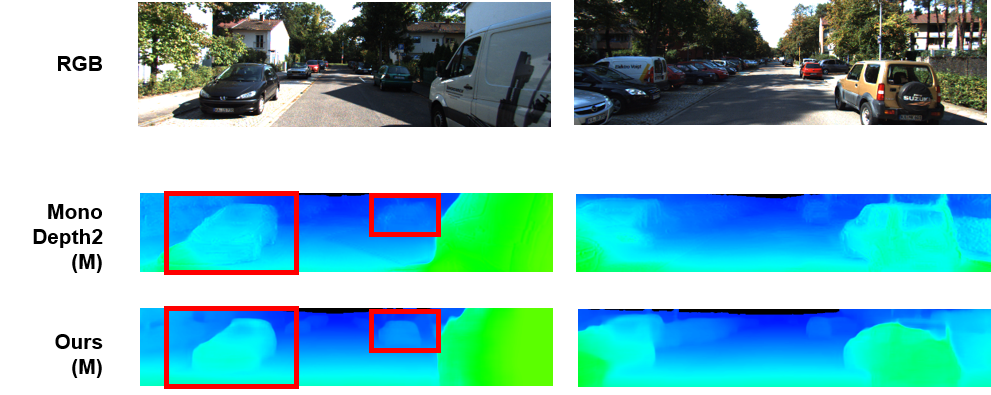}}
    \caption[Comparison of depth estimation results of Monodepth2 \cite{godard2019digging} and Ours (M)]{Comparison of depth estimation results of Monodepth2 \cite{godard2019digging} and Ours (M)
    } \label{Depth Comparison1-fig}
\end{figure}

\begin{figure}[t]
    \centerline{\includegraphics[width=8.5cm]{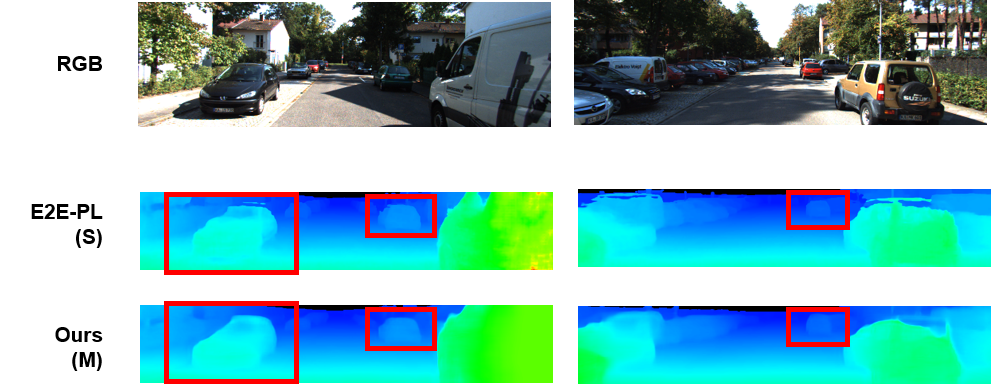}}
    \caption[Comparison of depth estimation results of E2E-PL \cite{Qian_2020_CVPR} and Ours(M).]{Comparison of depth estimation results of E2E-PL \cite{Qian_2020_CVPR} and Ours (M).
    } \label{Depth Comparison2-fig}
\end{figure}

\begin{table*}[t]
\renewcommand{\arraystretch}{1.2}
\caption[손실 모드에 따른 깊이 추정 성능]{Depth estimation performance according to loss types and with or without Scaled Depth.}
\label{tb:ablation depth}
\begin{center}
%\begin{sc}
\begin{tabular}{|c|c|c|cccc|ccc|}
\hline
\textbf{Method} & \textbf{Train} & \textbf{Scaled Depth}& \bm{$Abs Rel$} & \bm{$Sq Rel$} & \bm{$RMSE$}& \bm{$RMSE_{log}$} & \bm{$\delta < 1.25$} & \bm{$\delta < 1.25^2$} & \bm{$\delta < 1.25^3$}\\
\hline\hhline{|=|=|=|====|===|}
MD2& M &-& 0.234  &2.101 &8.267& 0.400& 0.583&  0.787&  0.891 \\
\hline
MD2+PIXOR$^*$+E2E& M &no& 0.997 &16.475 &20.569& 6.091&0.000&0.000&0.000 \\
\hline
MD2+PIXOR$^*$+E2E& M &yes& 0.313 &4.906 &10.418& 0.361&0.602&0.817&0.919 \\
MD2+PIXOR$^*$+E2E& MD &yes& 0.111 & 0.780  &4.503&0.191 &0.868 &0.958&0.984\\ 
MD2+PIXOR$^*$+E2E& D &yes&\textbf{0.107}  & \textbf{0.736} &\textbf{4.415}&\textbf{0.186} &\textbf{0.872} &\textbf{0.960}&\textbf{0.984}\\
\hline
\end{tabular}
%\end{sc}
\end{center}
\end{table*}

\begin{table*}[t]
\renewcommand{\arraystretch}{1.2}
\caption[3D object detection result according to end-to-end learning and loss types.]{3D object detection result according to end-to-end learning and loss types.}
\label{tb:ablation AP}
\begin{center}
\begin{tabular}{|c|c|ccc|ccc|}
\hline
\multicolumn{1}{|c}{\multirow{2}[2]{*}{\textbf{Method}}} & \multicolumn{1}{|c}{\multirow{2}[2]{*}{\textbf{Train}}} & \multicolumn{3}{|c}{{$\textbf{AP}_\textbf{BEV}$} (\%), \textbf{IoU=0.5}} & \multicolumn{3}{|c|}{{$\textbf{AP}_\textbf{BEV}$} (\%), \textbf{IoU=0.7}}\\
%\multirow{2}{*}{\textbf{Method}} & \multirow{2}{*}{\textbf{Train}} &
\cline{3-5}
\cline{6-8}
& & Easy & Moderate & Hard & Easy & Moderate & Hard\\
\hline\hhline{|=|=|===|===|}
MD2+PIXOR$^*$ & M & 8.73 & 6.93 & 5.36 & 4.55 & 4.55 & 4.55 \\
    MD2+PIXOR$^*$+E2E& M & 64.35& 50.39 & 44.03& 28.25& 23.19 & 21.01\\
    MD2+PIXOR$^*$+E2E& MD &69.85 & 58.83 & 53.72 & 37.02 &31.00 &\textbf{29.26}\\
    MD2+PIXOR$^*$+E2E& D &\textbf{70.86}& \textbf{60.23} &\textbf{54.54}&  \textbf{37.42}&\textbf{31.05}& 29.13\\
\hline
\end{tabular}
\end{center}
\end{table*}  

\textbf{Comparative Study.}
The depth estimation results of E2E-PL and our method and the 3D object detection results are visualized in Figure \ref{overview-fig}. In the case of E2E-PL, it is outputting much more predictions than 3D object labels. On the other hand, our proposed method outputs the number of predictions that closely match the label, so the number of false positives is significantly less than that of E2E-PL.

In Fig. \ref{Depth Comparison1-fig}, we qualitatively compare the depth estimation results of Monodepth2 \cite{godard2019digging} and our proposed method. Monodepth2 is not able to accurately detect distant objects or mistook image features such as shadows for depth in many cases. In particular, in the first picture, the shadow of the roof of the house reflected on the floor still exists at the depth predicted by Monodepth2. However, our method is accurately predicting the depth.

Also, in Fig. \ref{Depth Comparison2-fig}, we compare the depth estimation results of E2E-PL \cite{Qian_2020_CVPR} with our method (M). Since E2E-PL uses stereo images and 3D LiDAR, its performance is inevitably superior to that of monocular images alone, but our algorithm nevertheless shows similar depth estimation performance to E2E-PL.

We quantitatively compare the performance of our proposed method with existing monocular image-based methods. Because we used PIXOR as the 3D object detection network, we could only evaluate $AP_{BEV}$ (\%), and the results can be seen in Table \ref{tb:result_comparative}. \iffalse PL-MONO \cite{Wang_2019_CVPR}, Mono PL \cite{weng2019monocular} used DORN \cite{fu2018deep} as the depth estimation network, while DORN used LiDAR for pre-training.\fi As a result of comparing algorithms using only monocular images, our result, which is trained with unsupervised loss (M), outperformed most difficulty levels of the KITTI 3D validation set and showed the best performance. As a result of learning with supervised loss (D), our proposed method shows the SOTA performance in moderate and hard difficulty levels of car category in the KITTI 3D validation set.

\section{Conclusion}

In this paper, we proposed a network that learns with self-supervised loss using only monocular image sequences. The proposed monocular depth estimation network with a consistent scale is trained through end-to-end learning. The proposed network improved the performance of the monocular-based method in detecting 3D objects and narrowed the performance gap with the stereo-based method.

Despite these advances, the monocular image-based method is still less accurate than the LiDAR-based method by about 10\% $AP_{BEV}$. Also, as we use reduced image size to 640$\times$192 when inputting to the network, there is a problem in that a distant object is expressed in a small size in the image, making it more difficult to detect. In addition, if there is a moving object in the image sequence, an inaccurate unsupervised loss is calculated. In the future, we will study further how to isolate the pixels corresponding to moving objects and treat them differently by generating binary masks and attention map.
%%%%%%%%%%%%%%%%%%%%%%%%%%%%%%%%%%%%%%%%%%%%%%%%%%%%%%%%%%%%%%%%%%%%%%%%%%%%%%%%

\bibliographystyle{ieeetr}
\bibliography{reference}

\end{document}